\pdfoutput=1

\documentclass[11pt]{article}

\usepackage{acl}

\usepackage{times}
\usepackage{latexsym}
\usepackage{microtype}
\usepackage{adjustbox}
\usepackage{booktabs}
\usepackage{footnote}
\usepackage{subcaption}
\usepackage{url}
\usepackage{soul}

\usepackage[T1]{fontenc}

\usepackage[utf8]{inputenc}

\usepackage{microtype}
\usepackage{multirow}

\title{Assessing the Portability of Parameter Matrices Trained by Parameter-Efficient Finetuning Methods}

\author{Mohammed Sabry \\
  ADAPT/DCU, Dublin, Ireland \\
  \texttt{mohammed.sabry@adaptcentre.ie}\And
  Anya Belz \\
  ADAPT/DCU, Dublin, Ireland\\
  \texttt{anya.belz@adaptcentre.ie} \\}

\begin{document}
\maketitle

\begin{abstract}
As the cost of training ever larger language models has grown, so has the interest in reusing previously learnt knowledge. Transfer learning methods have shown how reusing non-task-specific knowledge can help in subsequent task-specific learning.
In this paper, we investigate the inverse: porting whole functional modules that encode task-specific knowledge from one model to another. 
We designed a study comprising 1,440 training/testing runs to  test the portability of modules trained by parameter-efficient finetuning (PEFT) techniques, using sentiment analysis as an example task. We test portability in a wide range of scenarios, involving different PEFT techniques and different pretrained host models, among other dimensions. We compare the performance of ported modules with that of equivalent modules trained (i) from scratch, and (ii) from parameters sampled from the same distribution as the ported module.
We find that the ported modules far outperform the two alternatives tested, but that there are interesting performance differences between the four PEFT techniques.
We conclude that task-specific knowledge in the form of structurally modular sets of parameters as produced by PEFT techniques is highly portable, but that degree of success depends on type of PEFT and on differences between originating and receiving pretrained models.
\end{abstract}

\section{Introduction and Related Work}

Given the increasing costs of training and running neural models \cite{strubell-etal-2019-energy}, the interest in finding methods to reduce these costs is growing. Reusability of previously learned knowledge is one very promising avenue to pursue, in particular if this were possible in plug-and-playable form. 

Methods that come under the broad heading of transfer learning have shown for some time that general, non-task-specific knowledge transferred from one learning scenario to another can help speed up task-specific learning in the latter. Well established techniques such as word and word-sequence embeddings, and pretraining plus finetuning are examples, as is adaptation from one domain to another \cite{guo2022domain}, one language to another \cite{conneau-etal-2020-unsupervised}, or one task to another \cite{ruder-etal-2019-transfer}. What  these approaches have in common is that they aim to extract general, or at least non-task-specific, knowledge while discarding the task-specific knowledge. 

Reusability could be radically extended if it were possible to reuse both generic and different types of task-specific knowledge, especially if these could be recombined with some degree of freedom. For this to be possible, the knowledge would have to be contained in structurally and functionally modular, or {\textbf{portable}}, (sub)networks. Some research has explored model compression \cite{jiang-etal-2023-pruning} which can be seen as attempting to extract modules with desired functionality. Other work has looked at identifying subnetworks with given functionality \cite{csordás2021neural}, but none has to our knowledge successfully demonstrated \textbf{{portability}} of task-specific modules. 

Parameter efficient finetuning (PEFT) techniques such as Adapters \cite{pmlr-v97-houlsby19a}, Prefix Tuning \cite{li-liang-2021-prefix}, Compacters \cite{NEURIPS2021_081be9fd}, and LoRA \cite{DBLP:journals/corr/abs-2106-09685}, train sets of parameters that have been shown to be \textit{structurally} modular \cite{sabry2023peftref}, in the sense that they form separate parameter sets that interact with their host model via dedicated interfaces. However, it is currently unclear if PEFT modules  are also \textit{functionally} modular. One important marker of functional modularity is \textbf{encapsulation}, i.e.\ the degree to which a (structural) module performs dedicated functions that are separate from functionality elsewhere in the system. Encapsulation is a precondition for {portability} which would be an important step in the direction of plug-and-playable neural components, potentially capable of achieving substantial reductions in training time and resources, and increased reusability in neural system development \cite{schmidt1998modularity, kingetsu2021neural, bhattacharya2022portability, pfeiffer2023modular}.

Modularity (without porting) has been explored in the context of Adapters for multi-task cross-lingual transfer \cite{pfeiffer-etal-2020-mad}. Cross-task transferability (in unchanged PEFT-tuned models) has also started to be explored very recently, e.g.\ in conjunction with prompt tuning \cite{su-etal-2022-transferability,vu-etal-2022-spot}.  \citet{Ding2023} extended this to other PEFT techniques, showing that PEFT-tuned models maintain performance on closely related tasks, but not on less closely related tasks. 

In this focused contribution, we assess something more challenging: whether PEFT techniques, specifically, create modules that encode task-specific knowledge that is portable to new models. We start with an overview of our study (Section \ref{peft_port_overview}) and the experimental set-up (Section~\ref{sec:peft-port-exp-setup}). We then present the results (Section~\ref{sec:peft_port_results}), and conclude with discussion and findings (Section~\ref{conclusion}).

\section{Study Overview}\label{peft_port_overview}

Our  goal in the present study is to investigate the degree to which the knowledge encoded in the parameter matrices that result from PEFT tuning (which we call \textbf{PEFT modules}) is portable. More specifically, the degree to which such knowledge is portable between different models under different conditions.

The  study is designed to test the portability of  modules trained by different PEFT techniques from an \textbf{originating model} (in conjunction with which the module was trained), to a different \textbf{receiving model}; moreover to test it under different conditions, including different types and combinations of originating and receiving models, different numbers of learning steps during module training at the originating model end, and (b) module training at the receiving model end, as described in more detail in the next section.

\begin{table}
\centering\begin{small}
\setlength\tabcolsep{4pt} 
\renewcommand{\arraystretch}{1.15}
\begin{tabular}{llcc}
\toprule
Instruction-tuned & \multirow{2}{*}{Raw Model}   & \multirow{2}{*}{\#Params} & \multirow{2}{*}{Learning Steps} \\ 
 Model &    &  &  \\ 
 \midrule
Flan T5 base& T5 v1 base  & 250M   & 84k\\
Flan T5 large   & T5 v1 large & 780M   & 64k\\
\bottomrule
\end{tabular}
\vspace{-0.1cm}
\caption{Pretrained models used,  raw/instruction-tuned variants, number of parameters and number of learning steps in instruction tuning \cite{chung2022scaling}.}
\vspace{-0.45cm}

\label{tab:instructionmodels}   
\end{small}
\end{table}

\begin{table}
\centering
\begin{small}
\setlength\tabcolsep{2pt} 
\renewcommand{\arraystretch}{1.15}
\begin{tabular}{lccccc}
\toprule
\multirow{2}{*}{PEFT}  & Archit. & \multirow{2}{*}{Repeats} & \multirow{2}{*}{Insertion}  & \multirow{2}{*}{Workspace}  \\ 
      & (MLP) &  &   &   \\ 
\midrule
Prefix Tun.\ &  Non-lin.\ & All layers  & Parallel  & Attn keys/values \\

LoRA  & Linear   & All layers  & Parallel & Attn query/val.\ \\

Adapter  &  Non-lin.\ & All layers  & Sequential  & FFN, Attn block\\

Compacter     &  Non-lin.\ & All layers  & Sequential  & FFN, Attn block     \\

\bottomrule
\end{tabular}
\end{small}
\vspace{-0.1cm}
\caption{PEFT techniques used in experiments, alongside  structural properties as per  \citet{sabry2023peftref}.} 
\vspace{-0.4cm}

\label{tab:peftport_properties_limitation}
\end{table}

\section{Experimental Set-Up}\label{sec:peft-port-exp-setup}

Put simply, if the knowledge encapsulated in PEFT modules is portable to new models, then plugging a pretrained PEFT module into a new model
will result in superior performance for the same number of post-porting learning steps than a randomly initialised PEFT module.

More strictly, if it really is the knowledge encapsulated by the pretrained PEFT module that leads to the superior performance rather than simply starting training off in a statistically advantageous point in the search space, then initialisation with parameters sampled from the same distribution (with the same mean and variance) will result in worse performance.

To establish whether these are the case is the purpose of the present study. In it we performed experiments as per the following experimental grid: (i) four combinations of originating and receiving models, (ii) sentiment classification as the example NLP task, (iii) four PEFT techniques, (iv) same vs.\ different datasets on originating and receiving sides, (v) two importing scenarios (exact parameters vs.\ sampled from same distribution), (vi) two different numbers of learning steps in the pre-porting training of modules, (vii) three different numbers of learning steps in post-porting training (module adaptation to the receiving model environment), and (viii) three different random seeds. This grid corresponds to 1,152 experiments; we added 288 experiments for training from scratch without importing the pretrained PEFT module (where there is no pre-porting training, and no importing scenarios), making it a total of 1,440 experiments.

\begin{table}[t!]
\setlength\tabcolsep{3pt} 
\renewcommand{\arraystretch}{1.15}
\begin{small}
\begin{subtable}{0.45\textwidth}
\begin{tabular}{lp{1.7cm}p{1.7cm}p{1.7cm}}
\toprule
& \multicolumn{3}{c}{\textit{Mean Accuracy (Variance)}}\\
\multicolumn{1}{l}{\textbf{PEFT}} & \multicolumn{1}{c}{\textbf{Ported}} & \multicolumn{1}{c}{\textbf{Sampled}} & \multicolumn{1}{c}{\textbf{From scratch}} \\ 
\midrule
Adapter  & 0.895 \begin{small}(0.001)\end{small}  & 0.777 \begin{small}(0.031)\end{small} &     0.765 \begin{small}(0.033)\end{small} \\
Compacter& 0.661 \begin{small}(0.037)\end{small}  & 0.478 \begin{small}(0.140)\end{small} &  0.477 \begin{small}(0.140)\end{small}\\
LoRA     & 0.600 \begin{small}(0.148)\end{small}  & 0.480 \begin{small}(0.189)\end{small} &    0.544 \begin{small}(0.166)\end{small}   \\
Prefix Tuning& 0.751 \begin{small}(0.010)\end{small}  &  0.692 \begin{small}(0.021)\end{small} &  0.685 \begin{small}(0.032)\end{small}\\ \bottomrule
\end{tabular}
\caption{Task tuning on originating side and adaptation tuning on receiving side use \textit{same} dataset (Rotten Tomatoes).}
\label{tab:same_dataset_ablation_instruction_1}
\end{subtable}

\vspace{5pt} 

\begin{subtable}{0.45\textwidth}
\begin{tabular}{lp{1.75cm}p{1.75cm}p{1.75cm}}
\midrule
Adapter & 0.930 \begin{small}(0.005)\end{small}  & 0.797 \begin{small}(0.040)\end{small} &     0.785 \begin{small}(0.041)\end{small} \\
Compacter & 0.681 \begin{small}(0.037)\end{small}  & 0.493 \begin{small}(0.147)\end{small} &  0.481 \begin{small}(0.147)\end{small}\\
LoRA   & 0.629 \begin{small}(0.157)\end{small}  & 0.502 \begin{small}(0.205)\end{small} &    0.561 \begin{small}(0.179)\end{small}   \\
Prefix Tuning  & 0.829 \begin{small}(0.005)\end{small}  &  0.734 \begin{small}(0.027)\end{small} &  0.743 \begin{small}(0.020)\end{small}\\ 
\bottomrule
\end{tabular}
\caption{Task tuning on originating side and adaptation tuning on receiving side use \textit{different} datasets (Rotten Tomatoes and SST-2, respectively).}
\label{tab:different_dataset_ablation_instruction_1}
\vspace{-0.1cm}
\end{subtable}
\end{small}

\caption{Mean Accuracy (Variance) for Ported, Sampled and From-scratch scenarios, broken down into results for \textit{same}/\textit{different} pre-porting and post-porting datasets.}
\label{tab:combined_ablation_results_1}
\vspace{-0.3cm}

\end{table}

\textbf{Pretrained models used (i above)}: We selected four different versions of the open-source T5 model \cite{10.5555/3455716.3455856}, as shown in Table~\ref{tab:instructionmodels}: T5 v1\footnote{\url{https://huggingface.co/docs/transformers/model_doc/t5v1.1}} raw\footnote{`Raw' refers to unaltered pretrained models without instruction-tuning; `base' refers to the smallest-size model.} models of two different sizes (250M, 780M) and their instruction-tuned Flan equivalents.\footnote{\url{https://huggingface.co/docs/transformers/model_doc/flan-t5}}
This selection gives us good coverage in terms of model size and types of knowledge in  pretrained models (raw language model vs.\ instruction tuned).


\textbf{Datasets (ii):} Our example NLP task is sentiment analysis, and we used two English datasets, namely SST-2\footnote{\url{https://huggingface.co/datasets/sst2} (train: 60.6K, val: 6.7K, test: 872) \cite{socher-etal-2013-recursive}} and Rotten Tomatoes,\footnote{\url{https://huggingface.co/datasets/rotten_tomatoes} (train: 8.53k, val: 1.07k, test: 1.07k) \cite{pang-lee-2005-seeing}} with Accuracy as the performance measure. The task was construed as sequence prediction, i.e.\ the input is provided as the prompt directly without task descriptions or prefixes, and the sequence continuation generated by the model should be the desired output (here, the sentiment label: `great' or `terrible'). We opted for this setup to ensure a level playing field for raw and instruction-tuned models. It avoids granting the latter an unfair advantage that could result from explicit task descriptions. The (raw) T5 v1 models (Table~\ref{tab:instructionmodels}) were pretrained exclusively on the Google C4 crawled dataset \cite{10.5555/3455716.3455856}, with no supervised training, so using a task prefix during single-task fine-tuning does not confer a real advantage, as it does, in contrast, for instruction-tuned models.\footnote{ \url{https://huggingface.co/docs/transformers/model_doc/t5v1.1}}

\textbf{PEFT techniques (iii):} The four PEFT techniques we tested\footnote{Implementations from OpenDelta, an open-source library for parameter-efficient finetuning: \url{https://github.com/thunlp/OpenDelta/tree/main}.} were Prefix Tuning \cite{li-liang-2021-prefix}, LoRA \cite{DBLP:journals/corr/abs-2106-09685}, Adapter \cite{pmlr-v97-houlsby19a}, and Compacter \cite{NEURIPS2021_081be9fd} each representing a different approach to parameter-efficient finetuning with different associated degrees of structural and functional modularity in resulting PEFT modules. An overview of their structural properties, in terms of the PEFT-Ref typology \cite{sabry2023peftref}, is provided in Table~\ref{tab:peftport_properties_limitation}: architecture (Column~2), number of insertions across transformer layers (Column~3), in-parallel versus sequential insertion (Column~4), and which parameters in the transformer layer architecture they interact with (their `workspace', Column~5).

\begin{figure*}[t] 
    \centering
    \begin{tabular}{c}
       \begin{small} S a m e \hspace{0.2cm}d a t a s e t s\end{small}\\
    \end{tabular}
    
    \begin{subfigure}{0.49\textwidth} 
        \centering
        \includegraphics[width=\textwidth]{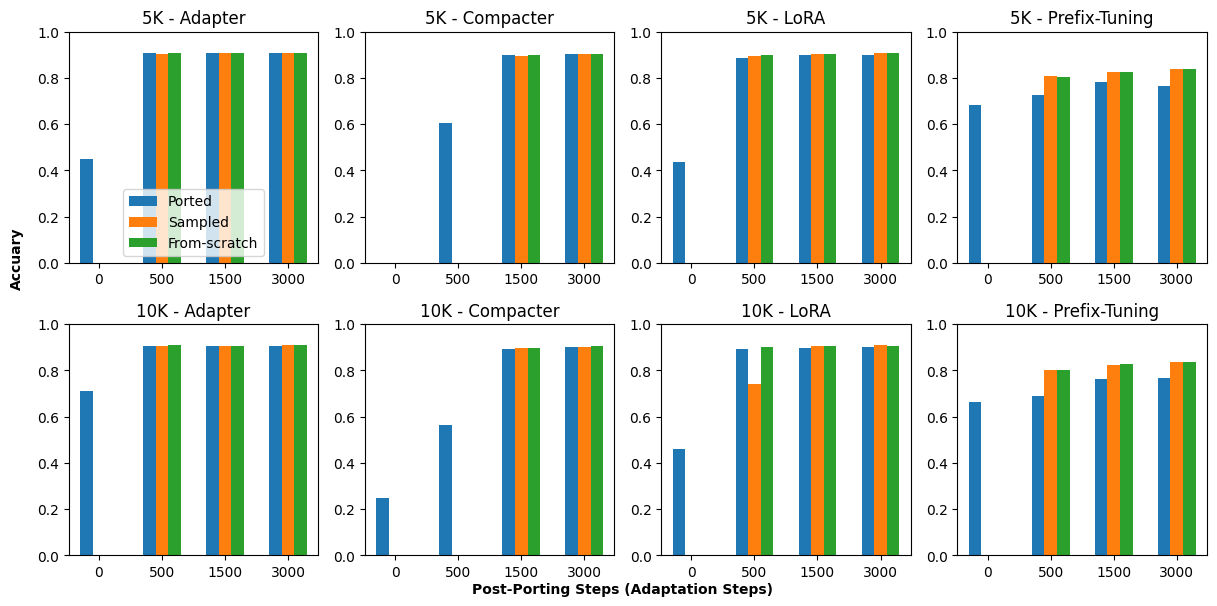}
        \caption{Porting direction: raw $\rightarrow$ instruction-tuned}
    \end{subfigure}
    \hfill
    \begin{subfigure}{0.49\textwidth} 
        \centering
        \includegraphics[width=\textwidth]{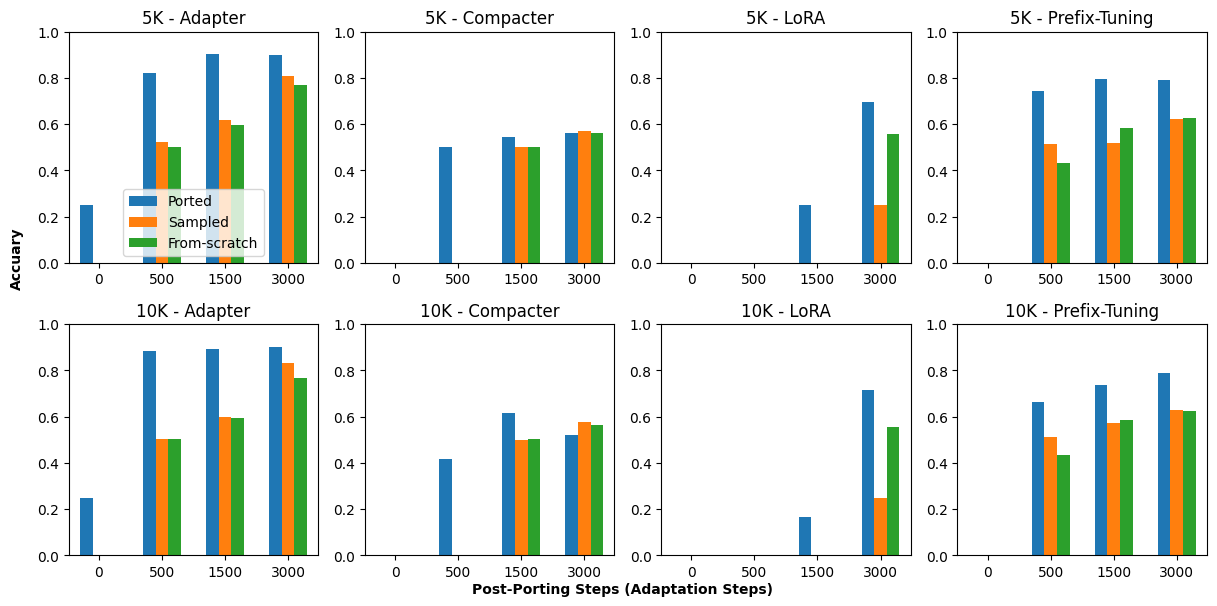}
        \caption{Porting direction: instruction-tuned $\rightarrow$ raw}
    \end{subfigure}
    
    \begin{tabular}{c}
      \begin{small} D i f f e r e n t \hspace{0.2cm}d a t a s e t s\end{small}\\
    \end{tabular}
    
    \begin{subfigure}{0.49\textwidth} 
        \centering
        \includegraphics[width=\textwidth]{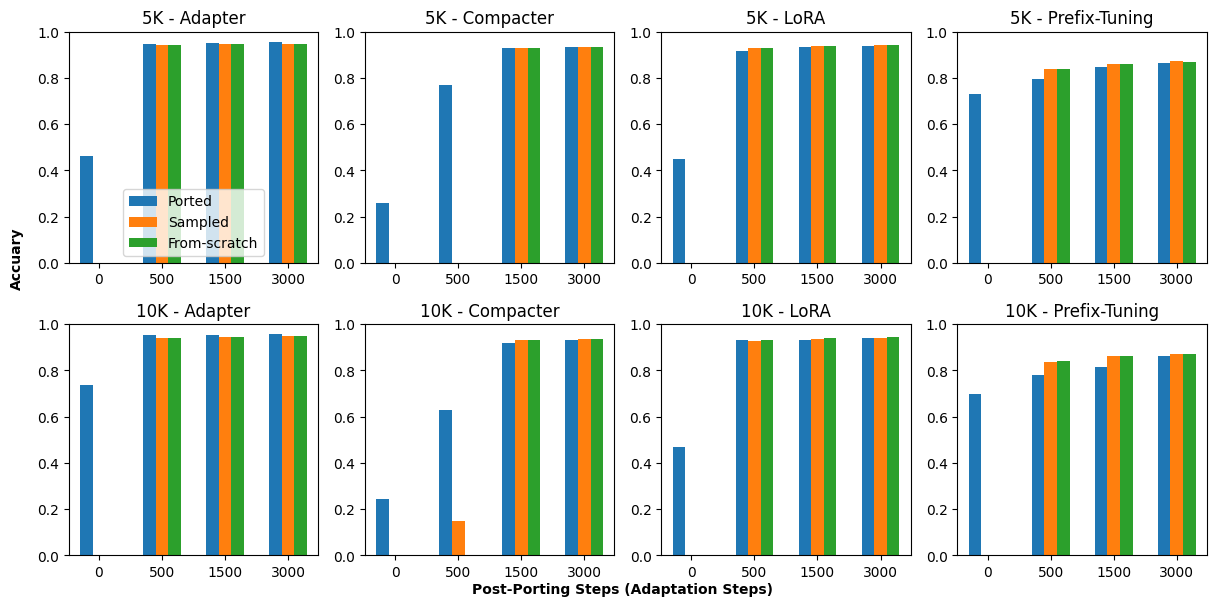}
        \caption{Porting direction: raw $\rightarrow$ instruction-tuned}
    \end{subfigure}
    \hfill
    \begin{subfigure}{0.49\textwidth} 
        \centering
        \includegraphics[width=\textwidth]{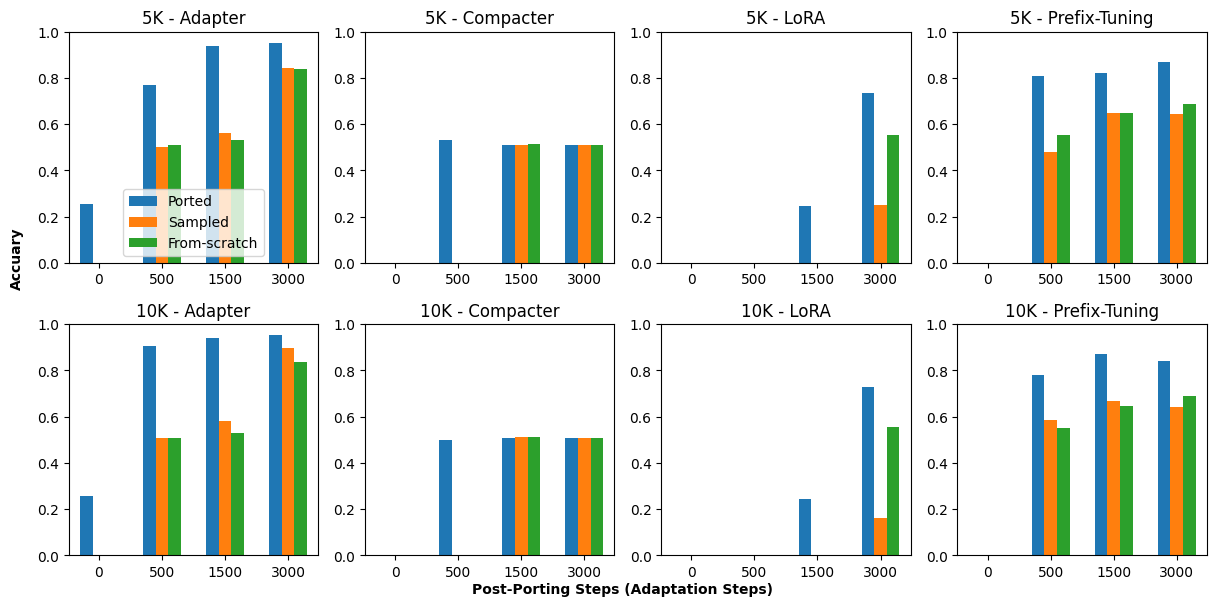}
        \caption{Porting direction: instruction-tuned $\rightarrow$ raw}
    \end{subfigure}
    
    \caption{Each bar chart shows average accuracy over three random seeds and two pairs of originating and receiving models for one PEFT technique (e.g. Adapter), one porting direction (e.g. raw $\rightarrow$ instruction-tuned), and one number of pre-porting training learning steps (e.g. 5K). Y-axis in each chart is Accuracy, X-axis is the number of post-porting adaptation learning steps (500, 1.5K and 3K), blue=ported, orange=sampled, and green=random parameters.}\label{fig:finegrained-results}
\end{figure*}

\textbf{Combinations of pre-porting and post-porting datasets (iv):} The pre-porting dataset is the one used to PEFT-tune the module (i.e.\ before it is exported). The post-porting dataset is the one used in further tuning an imported PEFT module within its new environment.
We compare (a) using the same dataset (Rotten Tomatoes) in post-porting tuning and testing as was used in pre-porting tuning, and (b) using different datasets (Rotten Tomatoes on the pre-porting side, and SST-2 on the post-porting side).

\textbf{Importing scenarios (v, additionally from-scratch tuning):} In this experimental dimension we tested three alternatives, namely (i) importing PEFT module parameters exactly as they are at the end of (pre-porting) PEFT-tuning, (ii) sampling new parameters from the same (normal) distribution, i.e.\ with the same mean and variance, and (iii) initialising parameters randomly using their PEFT default initialisation techniques\footnote{LoRA initialises all parameters with zero, Adapter uses normal distribution with mean $0$ and standard deviation $0.01$, Compacter uses Glorot uniform \cite{pmlr-v9-glorot10a}, and Prefix-Tuning uses the default PyTorch uniform initialisation for linear layers, then tuning from scratch.}.

\textbf{{Pre-porting and post-porting learning steps (vi, vii):}} We tested two different numbers of learning steps for pre-porting PEFT tuning: 5K and 10K. 
On the post-porting side, we tested three different numbers of learning steps: 
0.5K, 1K and 3K.

For details of the \textbf{hyperparameters} we used with the different methods,
see Appendix~\ref{ported_exp_training_details}.

\section{Results}\label{sec:peft_port_results}

The first two sets of results we present (in Tables~\ref{tab:same_dataset_ablation_instruction_1} and~\ref{tab:different_dataset_ablation_instruction_1}) are the mean and variance of Accuracy scores over all of the following experimental dimensions: \textit{i} (four pairs of originating and receiving models), \textit{vi} (two different numbers of pre-porting learning steps), \textit{vii} (three different numbers of post-porting learning steps), and \textit{viii} (three different random seeds). 
This provides a high-level perspective on the extent to which knowledge has been successfully ported on average for each of the four types of PEFT module, as compared to the corresponding sampled and from-scratch parameters. 
Table~\ref{tab:same_dataset_ablation_instruction_1} shows results when the same data set (Rotten Tomatoes) is used for PEFT tuning on the originating side and post-porting tuning and testing on the receiving side. Table~\ref{tab:different_dataset_ablation_instruction_1} shows results when different datasets are used (Rotten Tomatoes pre-porting and SST-2 post-porting).

We can very clearly see the substantial advantage that importing a pretrained PEFT module brings for all four PEFT techniques. Performance increases are similar across PEFT techniques and same/different datasets, but Compacter benefits the most, followed by Adapter, LoRA and Prefix-Tuning. As indicated in Table~\ref{tab:peftport_properties_limitation} (Column~5), LoRA and Prefix-Tuning interact with their host model by accessing weights, while Adapters and Compacters interact with representations. These structural differences may explain the observed portability variations, as weights can be viewed as the model fingerprint, making portability more challenging compared to representations, which can be shared among different models.

Figure~\ref{fig:finegrained-results} shows more finegrained results, for same datasets at the top (a and b), and different datasets at the bottom (c and d). Each half of the figure is further divided into porting from raw to instruction-tuned host models (left) and vice versa (right). More information in figure caption. 

Accuracy is remarkably similar for same vs.\ different pre-porting and post-porting datasets across the different scenarios. This implies that the knowledge acquired is dataset-agnostic. It is also very
stable across 5K vs 10K PEFT-tuning steps on the
originating side.

The porting direction makes a big difference. When porting from a raw host model to an instruction-tuned one (left side of Figure~\ref{fig:finegrained-results}), we see the following pattern. Remarkably, all PEFT techniques exhibit some degree of zero-shot portability, with ported modules achieving up to around 0.7 Accuracy straight out of the box, compared to 0 for both sampling and random parameters. From 500 post-porting learning steps onwards, performance evens out between ported, sampled and random parameters, and also plateaus out, for Adapter, LoRA and prefix-tuning. For Compacter, this happens at 1,000 steps.

When porting from an instruction-tuned host model to a raw one (right side of Figure~\ref{fig:finegrained-results}), we see different patterns. 
Only Adapters exhibit any zero-shot portability in this porting direction, albeit at much reduced Accuracy levels. However, here the performance with imported modules remains much higher than with sampled and random parameters across all learning steps; this is the case for all PEFT techniques except Compacters. 
In terms of overall best performance, only Adapters match the corresponding best performance in the other porting direction (raw to instruction-tuned) by 3,000 learning steps.
LoRA and Compacter perform much less well overall than Adapter and prefix-tuning in this porting direction.

The differences between the two porting directions may be in part due to differences in knowledge encoded in raw and instruction-tuned models. 
A PEFT module trained with a raw model as host has to acquire all task-specific knowledge (because a straightforward language model has none), making the knowledge encapsulated in the PEFT module more task-specific and more self-contained, explaining the good zero-shot post-porting performance observed. At the same time, the receiving host model, because instruction-tuned, already has relevant task-specific knowledge, explaining why ported, sampled and random variants perform on a par from 500 (1,000 for Compacter) post-porting learning steps onward. 

Conversely, a PEFT module trained with an instruction-tuned model as host only has to acquire task-specific knowledge not already present in the host, making the knowledge encapsulated in the resulting PEFT module less task-specific and less self-contained, explaining the mostly absent zero-shot post-porting performance observed. At the same time, the partial task-specific knowledge encoded in the imported parameter still bestows a substantial boost in a situation where the receiving host model is a raw  model with no task-specific knowledge, explaining why the ported modules outperform alternatives in all scenarios except for Compacters with different datasets.

The results reported here are for a comparatively easy task. In Appendix~\ref{appendix:additional_experiments}, we report preliminary results for similar experiments involving Natural Language Inference, a much more complex task, with the aim of confirming generalisation to more complex tasks. 

\section{Conclusion}\label{conclusion}

Our study shows, for the first time, that PEFT modules are structurally and functionally sufficiently modular to be portable from one host model to another. Remarkably, we observed pronounced zero-shot portability (with no post-porting adaptation tuning at all) for the best PEFT techniques. The performance that can be achieved in the model being ported to depends on the porting direction and PEFT technique used. Adapters appear to deliver the highest degree of portability overall across both directions.

Given the structural differences between the types of PEFT modules tested, our results point in an exciting direction: it may be possible to extrapolate from such results to design new PEFT techniques specifically optimised for portability. The structural properties of current PEFT techniques impose limits on the reusability of ported modules, e.g.\ requiring the receiving model to have the same hidden dimension and number of layers as the originating model. Addressing these limitations could pave the way for more versatile and widely portable PEFT modules.

We are currently epxloring these aspects further in extended portability tests, initially for a wider range of different tasks, and subsequently for other models and task construals. A particular focus in future work will be the efficiency savings that can be achieved through portable modules, including computational budgets required for different PEFT techniques to achieve satisfactory performance in ported modules.

\section*{Limitations}

Our findings should be interpreted within the context of the selected models, datasets, task formulation, and hyperparameters. Our choice of hyperparameters for PEFT techniques is informed by prior research, and our selection of learning steps is driven by the goal of achieving performance while staying within computational constraints. In particular, we demonstrate portability for sentiment analysis, with some back up from the much more complex task of NLI.

\section*{Responsible Research Notes}

In the work reported here, we used open-source resources and datasets only. These are all used in exactly the way they were intended to be used, for scientific research.

We used two of the standard sentiment analysis datasets that have been widely used in the field. We did not ourselves check for personally identifiable information or offensive content in these datasets. We have provided references to the sources of the datasets used which provide information regarding data collection and processing steps.

As work that uses standard open source datasets and standard opensource models and parameter-efficient finetuning techniques with automatic evaluation, the present work can be considered low-risk in terms of ethical consideration. Working on parameter-efficient finetuning and reusability will hopefully contribute to more energy-conserving model training and usage.

\bibliography{anthology,custom}

\begin{thebibliography}{26}
\expandafter\ifx\csname natexlab\endcsname\relax\def\natexlab#1{#1}\fi

\bibitem[{Bhattacharya et~al.(2022)Bhattacharya, Calafiura, Childers, Dewing, Dong, Gutsche, Habib, Ju, Kirby, Knoepfel, Kortelainen, Kwok, Leggett, Lin, Pascuzzi, Strelchenko, Viren, Yeo, and Yu}]{bhattacharya2022portability}
Meghna Bhattacharya, Paolo Calafiura, Taylor Childers, Mark Dewing, Zhihua Dong, Oliver Gutsche, Salman Habib, Xiangyang Ju, Michael Kirby, Kyle Knoepfel, Matti Kortelainen, Martin Kwok, Charles Leggett, Meifeng Lin, Vincent~R. Pascuzzi, Alexei Strelchenko, Brett Viren, Beomki Yeo, and Haiwang Yu. 2022.
\newblock \href {http://arxiv.org/abs/2203.09945} {Portability: A necessary approach for future scientific software}.

\bibitem[{Chung et~al.(2022)Chung, Hou, Longpre, Zoph, Tay, Fedus, Li, Wang, Dehghani, Brahma, Webson, Gu, Dai, Suzgun, Chen, Chowdhery, Castro-Ros, Pellat, Robinson, Valter, Narang, Mishra, Yu, Zhao, Huang, Dai, Yu, Petrov, Chi, Dean, Devlin, Roberts, Zhou, Le, and Wei}]{chung2022scaling}
Hyung~Won Chung, Le~Hou, Shayne Longpre, Barret Zoph, Yi~Tay, William Fedus, Yunxuan Li, Xuezhi Wang, Mostafa Dehghani, Siddhartha Brahma, Albert Webson, Shixiang~Shane Gu, Zhuyun Dai, Mirac Suzgun, Xinyun Chen, Aakanksha Chowdhery, Alex Castro-Ros, Marie Pellat, Kevin Robinson, Dasha Valter, Sharan Narang, Gaurav Mishra, Adams Yu, Vincent Zhao, Yanping Huang, Andrew Dai, Hongkun Yu, Slav Petrov, Ed~H. Chi, Jeff Dean, Jacob Devlin, Adam Roberts, Denny Zhou, Quoc~V. Le, and Jason Wei. 2022.
\newblock \href {http://arxiv.org/abs/2210.11416} {Scaling instruction-finetuned language models}.

\bibitem[{Conneau et~al.(2020)Conneau, Khandelwal, Goyal, Chaudhary, Wenzek, Guzm{\'a}n, Grave, Ott, Zettlemoyer, and Stoyanov}]{conneau-etal-2020-unsupervised}
Alexis Conneau, Kartikay Khandelwal, Naman Goyal, Vishrav Chaudhary, Guillaume Wenzek, Francisco Guzm{\'a}n, Edouard Grave, Myle Ott, Luke Zettlemoyer, and Veselin Stoyanov. 2020.
\newblock \href {https://doi.org/10.18653/v1/2020.acl-main.747} {Unsupervised cross-lingual representation learning at scale}.
\newblock In \emph{Proceedings of the 58th Annual Meeting of the Association for Computational Linguistics}, pages 8440--8451, Online. Association for Computational Linguistics.

\bibitem[{Csordás et~al.(2021)Csordás, van Steenkiste, and Schmidhuber}]{csordás2021neural}
Róbert Csordás, Sjoerd van Steenkiste, and Jürgen Schmidhuber. 2021.
\newblock \href {http://arxiv.org/abs/2010.02066} {Are neural nets modular? inspecting functional modularity through differentiable weight masks}.

\bibitem[{Ding et~al.(2023)Ding, Qin, Yang, Wei, Yang, Su, Hu, Chen, Chan, Chen, Yi, Zhao, Wang, Liu, Zheng, Chen, Liu, Tang, Li, and Sun}]{Ding2023}
Ning Ding, Yujia Qin, Guang Yang, Fuchao Wei, Zonghan Yang, Yusheng Su, Shengding Hu, Yulin Chen, Chi-Min Chan, Weize Chen, Jing Yi, Weilin Zhao, Xiaozhi Wang, Zhiyuan Liu, Hai-Tao Zheng, Jianfei Chen, Yang Liu, Jie Tang, Juanzi Li, and Maosong Sun. 2023.
\newblock \href {https://doi.org/10.1038/s42256-023-00626-4} {Parameter-efficient fine-tuning of large-scale pre-trained language models}.
\newblock \emph{Nature Machine Intelligence}, 5(3):220--235.

\bibitem[{Glorot and Bengio(2010)}]{pmlr-v9-glorot10a}
Xavier Glorot and Yoshua Bengio. 2010.
\newblock \href {https://proceedings.mlr.press/v9/glorot10a.html} {Understanding the difficulty of training deep feedforward neural networks}.
\newblock In \emph{Proceedings of the Thirteenth International Conference on Artificial Intelligence and Statistics}, volume~9 of \emph{Proceedings of Machine Learning Research}, pages 249--256, Chia Laguna Resort, Sardinia, Italy. PMLR.

\bibitem[{Guo and Yu(2022)}]{guo2022domain}
Xu~Guo and Han Yu. 2022.
\newblock \href {http://arxiv.org/abs/2211.03154} {On the domain adaptation and generalization of pretrained language models: A survey}.

\bibitem[{Houlsby et~al.(2019)Houlsby, Giurgiu, Jastrzebski, Morrone, De~Laroussilhe, Gesmundo, Attariyan, and Gelly}]{pmlr-v97-houlsby19a}
Neil Houlsby, Andrei Giurgiu, Stanislaw Jastrzebski, Bruna Morrone, Quentin De~Laroussilhe, Andrea Gesmundo, Mona Attariyan, and Sylvain Gelly. 2019.
\newblock \href {https://proceedings.mlr.press/v97/houlsby19a.html} {Parameter-efficient transfer learning for {NLP}}.
\newblock In \emph{Proceedings of the 36th International Conference on Machine Learning}, volume~97 of \emph{Proceedings of Machine Learning Research}, pages 2790--2799. PMLR.

\bibitem[{Hu et~al.(2021)Hu, Shen, Wallis, Allen{-}Zhu, Li, Wang, and Chen}]{DBLP:journals/corr/abs-2106-09685}
Edward~J. Hu, Yelong Shen, Phillip Wallis, Zeyuan Allen{-}Zhu, Yuanzhi Li, Shean Wang, and Weizhu Chen. 2021.
\newblock \href {http://arxiv.org/abs/2106.09685} {Lora: Low-rank adaptation of large language models}.
\newblock \emph{CoRR}, abs/2106.09685.

\bibitem[{Jiang et~al.(2023)Jiang, Wang, Zhuang, Xie, and Xia}]{jiang-etal-2023-pruning}
Ting Jiang, Deqing Wang, Fuzhen Zhuang, Ruobing Xie, and Feng Xia. 2023.
\newblock \href {https://doi.org/10.18653/v1/2023.acl-long.35} {Pruning pre-trained language models without fine-tuning}.
\newblock In \emph{Proceedings of the 61st Annual Meeting of the Association for Computational Linguistics (Volume 1: Long Papers)}, pages 594--605, Toronto, Canada. Association for Computational Linguistics.

\bibitem[{Karimi~Mahabadi et~al.(2021)Karimi~Mahabadi, Henderson, and Ruder}]{NEURIPS2021_081be9fd}
Rabeeh Karimi~Mahabadi, James Henderson, and Sebastian Ruder. 2021.
\newblock \href {https://proceedings.neurips.cc/paper_files/paper/2021/file/081be9fdff07f3bc808f935906ef70c0-Paper.pdf} {Compacter: Efficient low-rank hypercomplex adapter layers}.
\newblock In \emph{Advances in Neural Information Processing Systems}, volume~34, pages 1022--1035. Curran Associates, Inc.

\bibitem[{Kingetsu et~al.(2021)Kingetsu, Kobayashi, and Suzuki}]{kingetsu2021neural}
Hiroaki Kingetsu, Kenichi Kobayashi, and Taiji Suzuki. 2021.
\newblock \href {http://arxiv.org/abs/2112.13208} {Neural network module decomposition and recomposition}.

\bibitem[{Li and Liang(2021)}]{li-liang-2021-prefix}
Xiang~Lisa Li and Percy Liang. 2021.
\newblock \href {https://doi.org/10.18653/v1/2021.acl-long.353} {Prefix-tuning: Optimizing continuous prompts for generation}.
\newblock In \emph{Proceedings of the 59th Annual Meeting of the Association for Computational Linguistics and the 11th International Joint Conference on Natural Language Processing (Volume 1: Long Papers)}, pages 4582--4597, Online. Association for Computational Linguistics.

\bibitem[{Marelli et~al.(2014)Marelli, Menini, Baroni, Bentivogli, Bernardi, and Zamparelli}]{marelli-etal-2014-sick}
Marco Marelli, Stefano Menini, Marco Baroni, Luisa Bentivogli, Raffaella Bernardi, and Roberto Zamparelli. 2014.
\newblock \href {http://www.lrec-conf.org/proceedings/lrec2014/pdf/363_Paper.pdf} {A {SICK} cure for the evaluation of compositional distributional semantic models}.
\newblock In \emph{Proceedings of the Ninth International Conference on Language Resources and Evaluation ({LREC}'14)}, pages 216--223, Reykjavik, Iceland. European Language Resources Association (ELRA).

\bibitem[{Pang and Lee(2005)}]{pang-lee-2005-seeing}
Bo~Pang and Lillian Lee. 2005.
\newblock \href {https://doi.org/10.3115/1219840.1219855} {Seeing stars: Exploiting class relationships for sentiment categorization with respect to rating scales}.
\newblock In \emph{Proceedings of the 43rd Annual Meeting of the Association for Computational Linguistics ({ACL}{'}05)}, pages 115--124, Ann Arbor, Michigan. Association for Computational Linguistics.

\bibitem[{Pfeiffer et~al.(2023)Pfeiffer, Ruder, Vulić, and Ponti}]{pfeiffer2023modular}
Jonas Pfeiffer, Sebastian Ruder, Ivan Vulić, and Edoardo~Maria Ponti. 2023.
\newblock \href {http://arxiv.org/abs/2302.11529} {Modular deep learning}.

\bibitem[{Pfeiffer et~al.(2020)Pfeiffer, Vuli{\'c}, Gurevych, and Ruder}]{pfeiffer-etal-2020-mad}
Jonas Pfeiffer, Ivan Vuli{\'c}, Iryna Gurevych, and Sebastian Ruder. 2020.
\newblock \href {https://doi.org/10.18653/v1/2020.emnlp-main.617} {{MAD-X}: {A}n {A}dapter-{B}ased {F}ramework for {M}ulti-{T}ask {C}ross-{L}ingual {T}ransfer}.
\newblock In \emph{Proceedings of the 2020 Conference on Empirical Methods in Natural Language Processing (EMNLP)}, pages 7654--7673, Online. Association for Computational Linguistics.

\bibitem[{Raffel et~al.(2020)Raffel, Shazeer, Roberts, Lee, Narang, Matena, Zhou, Li, and Liu}]{10.5555/3455716.3455856}
Colin Raffel, Noam Shazeer, Adam Roberts, Katherine Lee, Sharan Narang, Michael Matena, Yanqi Zhou, Wei Li, and Peter~J. Liu. 2020.
\newblock Exploring the limits of transfer learning with a unified text-to-text transformer.
\newblock \emph{J. Mach. Learn. Res.}, 21(1).

\bibitem[{Ruder et~al.(2019)Ruder, Peters, Swayamdipta, and Wolf}]{ruder-etal-2019-transfer}
Sebastian Ruder, Matthew~E. Peters, Swabha Swayamdipta, and Thomas Wolf. 2019.
\newblock \href {https://doi.org/10.18653/v1/N19-5004} {Transfer learning in natural language processing}.
\newblock In \emph{Proceedings of the 2019 Conference of the North {A}merican Chapter of the Association for Computational Linguistics: Tutorials}, pages 15--18, Minneapolis, Minnesota. Association for Computational Linguistics.

\bibitem[{Sabry and Belz(2023)}]{sabry2023peftref}
Mohammed Sabry and Anya Belz. 2023.
\newblock \href {http://arxiv.org/abs/2304.12410} {Peft-ref: A modular reference architecture and typology for parameter-efficient finetuning techniques}.

\bibitem[{Schmidt and Bandar(1998)}]{schmidt1998modularity}
ALBRECHT Schmidt and ZUHAIR Bandar. 1998.
\newblock Modularity-a concept for new neural network architectures.
\newblock In \emph{Proc. IASTED International Conf. Computer Systems and Applications}, pages 26--29. Citeseer.

\bibitem[{Socher et~al.(2013)Socher, Perelygin, Wu, Chuang, Manning, Ng, and Potts}]{socher-etal-2013-recursive}
Richard Socher, Alex Perelygin, Jean Wu, Jason Chuang, Christopher~D. Manning, Andrew Ng, and Christopher Potts. 2013.
\newblock \href {https://aclanthology.org/D13-1170} {Recursive deep models for semantic compositionality over a sentiment treebank}.
\newblock In \emph{Proceedings of the 2013 Conference on Empirical Methods in Natural Language Processing}, pages 1631--1642, Seattle, Washington, USA. Association for Computational Linguistics.

\bibitem[{Strubell et~al.(2019)Strubell, Ganesh, and McCallum}]{strubell-etal-2019-energy}
Emma Strubell, Ananya Ganesh, and Andrew McCallum. 2019.
\newblock \href {https://doi.org/10.18653/v1/P19-1355} {Energy and policy considerations for deep learning in {NLP}}.
\newblock In \emph{Proceedings of the 57th Annual Meeting of the Association for Computational Linguistics}, pages 3645--3650, Florence, Italy. Association for Computational Linguistics.

\bibitem[{Su et~al.(2022)Su, Wang, Qin, Chan, Lin, Wang, Wen, Liu, Li, Li, Hou, Sun, and Zhou}]{su-etal-2022-transferability}
Yusheng Su, Xiaozhi Wang, Yujia Qin, Chi-Min Chan, Yankai Lin, Huadong Wang, Kaiyue Wen, Zhiyuan Liu, Peng Li, Juanzi Li, Lei Hou, Maosong Sun, and Jie Zhou. 2022.
\newblock \href {https://doi.org/10.18653/v1/2022.naacl-main.290} {On transferability of prompt tuning for natural language processing}.
\newblock In \emph{Proceedings of the 2022 Conference of the North American Chapter of the Association for Computational Linguistics: Human Language Technologies}, pages 3949--3969, Seattle, United States. Association for Computational Linguistics.

\bibitem[{Vu et~al.(2022)Vu, Lester, Constant, Al-Rfou{'}, and Cer}]{vu-etal-2022-spot}
Tu~Vu, Brian Lester, Noah Constant, Rami Al-Rfou{'}, and Daniel Cer. 2022.
\newblock \href {https://doi.org/10.18653/v1/2022.acl-long.346} {{SP}o{T}: Better frozen model adaptation through soft prompt transfer}.
\newblock In \emph{Proceedings of the 60th Annual Meeting of the Association for Computational Linguistics (Volume 1: Long Papers)}, pages 5039--5059, Dublin, Ireland. Association for Computational Linguistics.

\bibitem[{Williams et~al.(2018)Williams, Nangia, and Bowman}]{williams-etal-2018-broad}
Adina Williams, Nikita Nangia, and Samuel Bowman. 2018.
\newblock \href {https://doi.org/10.18653/v1/N18-1101} {A broad-coverage challenge corpus for sentence understanding through inference}.
\newblock In \emph{Proceedings of the 2018 Conference of the North {A}merican Chapter of the Association for Computational Linguistics: Human Language Technologies, Volume 1 (Long Papers)}, pages 1112--1122, New Orleans, Louisiana. Association for Computational Linguistics.

\end{thebibliography}
\bibliographystyle{acl_natbib}

\appendix

\section{Additional Experimental Details} \label{appendix:hyperparameters}

\subsection{Computational Resources}
Our experiments were conducted using a single NVIDIA A100 GPU with a memory capacity of 80GB.

\subsection{PEFT hyperparameters}\label{ported_exp_training_details}

Based on established practices in prior PEFT studies, we set the following hyperparameters for each technique:
\begin{itemize}
    \item \textbf{Adapters:}
Bottleneck dimension = $64$, 
Activation Function = $GeLU$.
\item \textbf{Compacter:}
Bottleneck dimension = $16$,
Activation Function = $GeLU$,
Hypercomplex division = $4$. 
No parameter-sharing between the Kronecker product reparameterised matrices.
\item \textbf{LoRA:}
Rank = $8$,
Alpha = $16$,
Dropout = $0.0$.
\item \textbf{Prefix Tuning:}
Number of tokens = $5$,
We employ a network comprising two linear layers with mid-dimensions = $512$.
The initial embedding dimension per token is set to $512$.
The activation function used in producing these tokens is $Tanh$.
The last layer is responsible for producing the desired token dimensions for the model.
\end{itemize}

\noindent In both  pre-porting and post-porting training, we utilised a learning rate of $1e-4$ with a linear decay scheduler. Additionally, we incorporated warm-up steps equivalent to $10\%$ of the total learning steps. Batch sizes were $4,096$ tokens in pre-porting training, and $2,048$ tokens in post-porting training. 

\section{Supplementary Experiments}\label{appendix:additional_experiments}

In order to confirm that portability of PEFT modules generalises beyond the tasks and datasets tested in this paper, more particularly to assess their performance in a more complex task, we conducted preliminary experiments on the task of Natural Language Inference (NLI), using the same experimental set-up (Section~\ref{sec:peft-port-exp-setup}).

We used two datasets, MNLI\footnote{\url{https://huggingface.co/datasets/SetFit/mnli}  (train: 393K, val: 9.8K, test: 9.8K) \cite{williams-etal-2018-broad}} and SICK,\footnote{\url{https://huggingface.co/datasets/sick} (train: 4.44K, val: 495, test: 4.91K) \cite{marelli-etal-2014-sick}} with the same task construal as for the experiments reported in the paper, namely providing the input directly as a prompt and interpreting the continuation generated as the output (here, NLI labels `neutral,' `entailment,' or `contradiction'). 

Again we used Accuracy as our performance metric. We tested for reduced ranges of pre-porting and post-porting learning steps, namely 5K pre-porting steps and 0.5K, 1K, and 3K post-porting steps. Moreover, we tested only the two most widely used PEFT techniques, Adapter and LoRA, with the same hyperparameters described in Appendix~\ref{appendix:hyperparameters}. 

In the same-dataset scenario, we used the MNLI dataset for pre-porting PEFT tuning and for post-porting adaptation tuning and evaluation. For the different-datasets scenario, we used MNLI on the pre-porting side and SICK on the post-porting side. We applied the same hyperparameters, as for the sentiment analysis experiments, except that the batch sizes for post-porting training were  
4,096 and 1,120 tokens for MNLI and SICK, reflecting different dataset characteristics. 

\begin{figure*}[!ht]

 \begin{tabular}{c}
      \hspace{6.75cm}\begin{small} S a m e \hspace{0.2cm} d a t a s e t s\end{small}\\
 \end{tabular}
 
    \begin{subfigure}{0.5\textwidth}
        \centering
        \includegraphics[width=\textwidth]{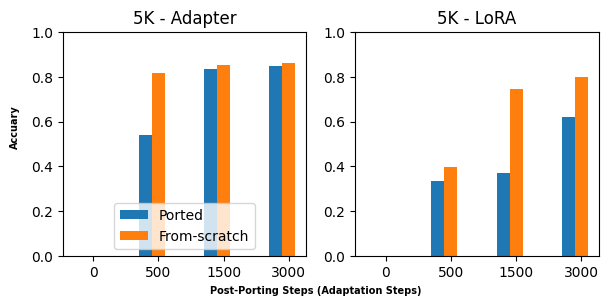}
        \caption{Porting direction: raw $\rightarrow$ instruction-tuned}
    \end{subfigure}
    \begin{subfigure}{0.5\textwidth}
        \centering
        \includegraphics[width=\textwidth]{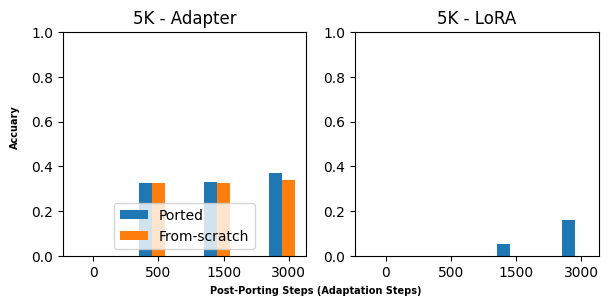}
        \caption{Porting direction: instruction-tuned $\rightarrow$ raw}
    \end{subfigure}
 
 \begin{tabular}{c}
      \hspace{6.25cm}\begin{small} D i f f e r e n t \hspace{0.2cm} d a t a s e t s\end{small}\\
 \end{tabular}

\begin{subfigure}{0.5\textwidth}
        \centering
        \includegraphics[width=\textwidth]{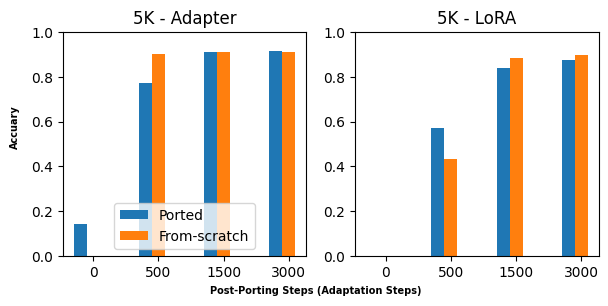}
        \caption{Porting direction: raw $\rightarrow$ instruction-tuned}
    \end{subfigure}
    \begin{subfigure}{0.5\textwidth}
        \centering
        \includegraphics[width=\textwidth]{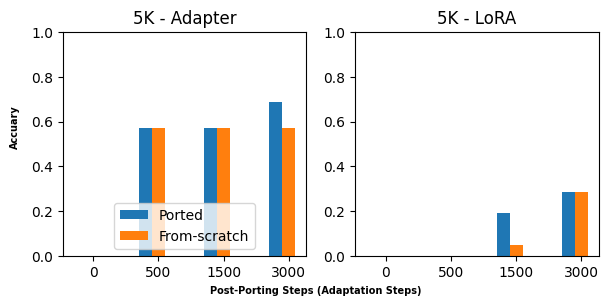}
        \caption{Porting direction: instruction-tuned $\rightarrow$ raw}
    \end{subfigure}
    \caption{Each bar chart shows average accuracy over three random seeds and two pairs of originating and receiving models for one PEFT technique (e.g.\ Adapter), one porting direction (e.g.\ raw $\rightarrow$ instruction-tuned), and one number of preporting training learning steps (e.g.\ 5K). Y-axis in each chart is Accuracy, x-axis is number of post-porting adaptation learning steps (500, 1.5K and 3K), blue=ported, orange=sampled, green=random parameters.}\label{fig:nli_finegrained-results}
    \vspace{-0.3cm}

\end{figure*}

The experimental set-up corresponds to a total of 288 experiments. 
The results (Figure~\ref{fig:nli_finegrained-results}) exhibit the same general patterns as described for the sentiment analysis tasks in Section~\ref{sec:peft_port_results}. However, we have so far tested only for two PEFT techniques, and only for what are very small numbers of pre-porting and post-porting learning steps for such a complex task, so the patterns are less clear. Nevertheless, Adapter and to a lesser degree LoRA successfully encapsulated and ported task-specific knowledge. The observed patterns align with our discussion of the influence of porting direction and PEFT structural properties in Section~\ref{sec:peft_port_results}. While these results indicate that PEFT portability generalises to more complex tasks, further research on a wider range of scenarios is needed. 

\end{document}